# Managing the Dynamics of a Harmonic Potential Field-Guided Robot in a Cluttered Environment

Ahmad A. Masoud
Electrical Engineering Department, KFUPM, P.O. Box 287, Dhaharan 31261, Saudi Arabia, masoud@kfupm.edu.sa

**Abstract-** This paper demonstrates the ability of the harmonic potential field (HPF) planning method to generate a well-behaved, constrained path for a robot with second order dynamics in a cluttered environment. It is shown that HPF-based controllers may be developed for holonomic as well as nonholonomic robots to effectively suppress the effect of inertial forces on the robot's trajectory while maintaining all the attractive features of a purely kinematic HPF planner. The capabilities of the suggested navigation controller are demonstrated using simulation results. Comparisons are also supplied with other approaches used for converting the guidance signal from a purely kinematic HPF planner into a navigation control signal.

## I. Introduction

A planner is an interface between an operator and a servo process whose function is to interpret the commands and constraints on the process behavior within the confines of the environment. The output of the planner is a context-sensitive, admissible, goal-oriented sequence of action instructions whose execution by the process actuators of motion produces a behavior that yields to the commands and constraints set by the operator. To function in this capacity the planner has to carry out several roles such as: changing the operator-centered format of the command and constraints on operation to a process-centered format. The planner must also act as a knowledge amplifier augmenting the partial information supplied by the operator to the minimum level needed by the process to execute the supplied task in the specified manner (figure-1).

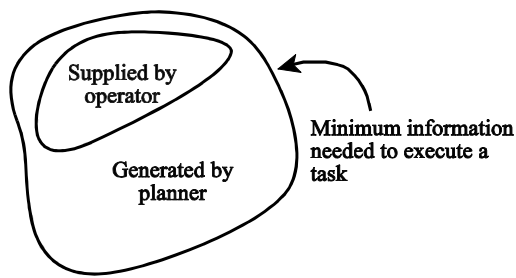

Figure-1: a successful planner should raise the level of information to the minimum needed for executing a task.

The above task is by no means simple, especially when servo processes with general dynamics are considered. Many of the practical aspects needed to construct planners that have a reasonable chance of success operating in a realistic environment are still open research problems. Understandably the literature abound with techniques and approaches for tackling this problem [1,2]. Despite the diversity of planning methods they may be divided into two classes: a class that separates a planner into two modules one called the high level controller (HLC) and the other is called the low level controller (LLC). The first is responsible for converting the command, constraints on process behavior, and environment feed into a desired behavior which the process must find a way to actualize if the task is to be accomplished (a know-what-to-do guidance signal). On the other hand, the second module determines what actions the process actuators of motion should release in order to actualize the desired behavior (a know-how-to-do control signal). Although this division of role in building planners is widely accepted by researchers in the area, it is believed to be a source of several problems. It is well-known in practice that processes using the HLC-LLC paradigm are relatively slow. Incompatibilities between the guidance and control signals could lead to unwanted artifacts in the behavior and undesirable control effort that consumes too much energy or put too much strain on the actuators. Jointly designing the guidance and control modules is expected to yield a simpler and more efficient planner compared to a design that treats the two modules separately.

Simultaneous consideration of the guidance and control signals in the design of a planner is a challenging task. While limited success was achieved in designing controllers that can incorporate simple avoidance regions with convex geometry in sate space [3,4], imposing general, nonconvex avoidance regions in the statespace of a dynamical system is difficult [5,6]. The task is further complicated when the state space constraints have to be implemented along with constraints in the control space as is the case with dynamical, nonholonomic systems.

Instead of using the relatively simple, two-tier approach to planner design or the excessively complex joint state space control space approach, an approach in the middle is adopted. Here the capabilities of a carefully selected planner that can only generate a guidance signal (i.e. deals only with the kinematic aspects of motion) are augmented to generate also the needed control signal. The guidance field from the kinematic planner is left unchanged. However, instead of the control component of the planner being designed to enforce strict compliance of motion with the guidance field, we only require that the control component strongly discourages motion from deviating from the course set by the guidance field. In its attempt to force compliance, the first approach injects too much energy into the system. This is expected to cause considerable transients in the response and an excessively high control effort. On the other hand, the passive nature of the suggested approach is highly unlikely to cause such problems.

As far as this work is concerned, the extremely rich variety of kinematic motion planners may be categorized in one of two classes: path tracking planners and goal seeking planners. A path tracking planner provides a sequence of guidance instructions that mark one and only one path from an initial state to a target state. If an unexpected event occur throwing the state away from the guidance path, it must find its way back to the path in order to proceed to the target. On the other hand, a goal seeking planner supplies a guidance instruction at every possible state the system may exist in. Therefore, a disruption caused by an influence external to the system will not cause a halt in the effort to drive the state closer to the target. For reasons that will become clearer in the sequel, goal seeking planners will be adopted in this work. In particular an efficient type of goal seeking planners known as harmonic potential field planners will be used.

This paper is organized as follows: section II provides a background of the HPF approach. Section III suggests a method for adapting the HPF method to deal with dynamic, holonomic systems. Section IV tackles the dynamic, nonholonomic, HPF case. Conclusions are in section V.

II. The HPF Approach - A Background

The micro element which an HPF planner utilizes for guiding the state of a system is a multi-dimensional vector attached to a specific point in state space. This element simply tells the system along which direction it should proceed if it is located at that state. A dense collective of theses vectors is induced using a surface (a potential field) along with a vector partial differential operator to fully cover the area of interest in the state space of the system (the workspace, $\Omega$). A group structure is then induced on this collective to generate a macro template with a structure encoding the guidance information the process needs to execute. The action selection mechanism the approach utilizes for generating the structure is in conformity with the artificial life (AL) method [7]. The HPF approach offers a solution to the local minima problem faced by the potential field approach Khatib suggested in [8]. It was simultaneously and independently proposed by several researchers [9-12] of whom the work of Sato in 1986 may be regarded as the first on the subject [13]. An HPF is generated using a Laplace boundary value problem (BVP) configured using a properly chosen set of boundary conditions. There are several settings one may use for a Laplce BVP (LBVP) in order to generate a navigation potential [14-16]. Each one of these settings possesses its own, distinct, topological properties [12]. An example is shown below of an LBVP that is configured using the homogeneous Neumann boundary conditions:

$$\nabla^2 V(X) \equiv 0 \qquad X \in \Omega \qquad (1)$$

subject to: $V(X_S) = 1$, $V(X_T) = 0$, and $\dfrac{\partial V}{\partial \mathbf{n}} = 0$ at $X = \Gamma$,

where $\Omega$ is the workspace, $\Gamma$ is its boundary, $\mathbf{n}$ is a unit vector normal to $\Gamma$, $X_s$ is the start point, and $X_T$ is the target point. The trajectory to the target (x(t)) is generated using the HPF-based, gradient dynamical system:

$$\dot{x} = -\nabla V(x) \qquad x(0) = x_0 \in \Omega \qquad (2)$$

The trajectory is guaranteed to:

1- $\lim_{t \to \infty} x(t) \to x_T$     2- $x(t) \in \Omega \qquad \forall t$

Harmonic functions have many useful properties[17] for motion planning. Most notably, a harmonic potential is also a Morse function and a general form of the navigation function suggested in [18]. The HPF approach may be configured to operate in a model-based and/or sensor-based mode. It can also be made to accommodate a variety of differential and state constraints [16]. It ought to be mentioned that the HPF approach is only a special case of a much larger class of planners called: evolutionary, pde-ode motion planners [14], figure-2.

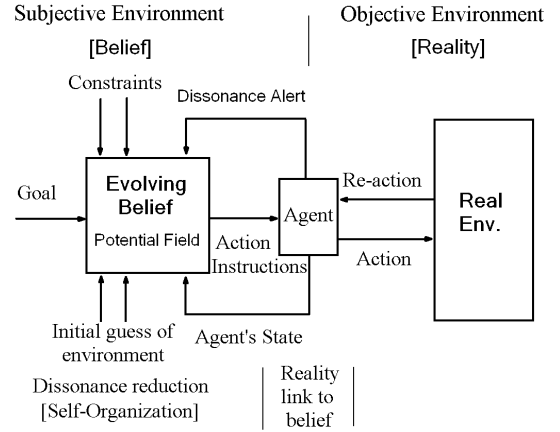

Figure-2: Block diagram of an evolutionary PDE-ODE planner.

Figures-3 show the guidance fields and paths generated by a special type of HPF planners [16] called nonlinear, anisotropic HPF planner (NAHPF). In addition to enforcing regional avoidance constraints, NAHPF planners can also enforce directional constraints.

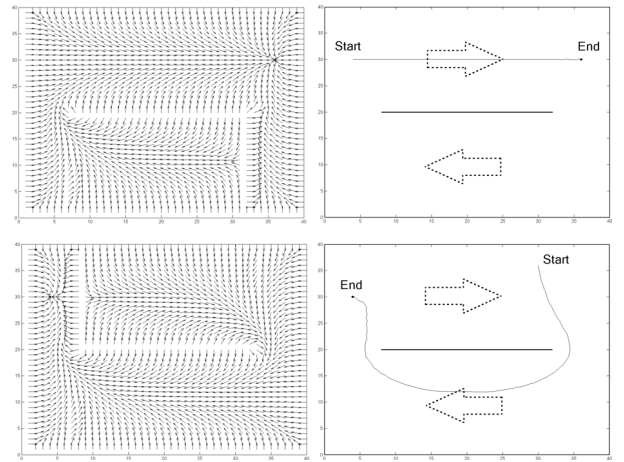

Figure-3: Output from a directional sensitive, kinematic, HPF planner.

The results in figure-3 are for a kinematic planner where the agent being guided is assumed to be a massless point. Assigning mass to the point robot totally changes the nature of the planning task. Here the planner must find the x and y force components which if applied to the point mass yield a trajectory similar to the one shown in the above figures. One way to generate the control signal is to treat the gradient guidance field as a driving force augmented with linear, viscous dampening force having a coefficient B [19]. For a 1 Kg mass, the system equation is:

$$\begin{bmatrix} \ddot{x} \\ \ddot{y} \end{bmatrix} = -\mathbf{B} \cdot \begin{bmatrix} \dot{x} \\ \dot{y} \end{bmatrix} - \begin{bmatrix} \partial V(x,y)/\partial x \\ \partial V(x,y)/\partial y \end{bmatrix} \qquad (3)$$

Unfortunately, the provably-correct properties of the kinematic planner can no longer be guaranteed. Figure-4 shows the kinodynamic planner with B=0.2. A mass of 1Kg was added. As can be seen the avoidance constraints failed and collision with the walls of the room did occur despite the fact that the initial speed is zero.

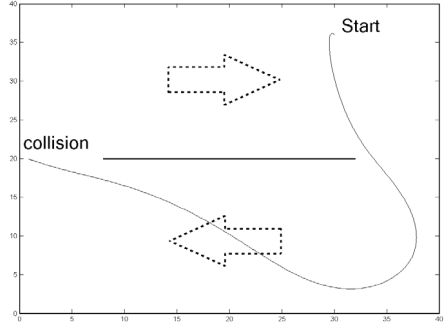

Figure-4: Inertial forces could lead to constraint violation.

### III. The Holonomic case

Increasing the coefficient of viscous dampening (B) may appear as the straight forward solution to the problem. Figure-5 shows that increasing B decreases the transients in the trajectory induced by the inertial forces. As demonstrated a high enough B has the ability to drive the spatial component of the dynamic trajectory arbitrarily close to the kinematic trajectory hence improving the chance of the planner to enforce the spatial constraints. The price to be paid for adopting such a simple solution is making the system impractically slow.

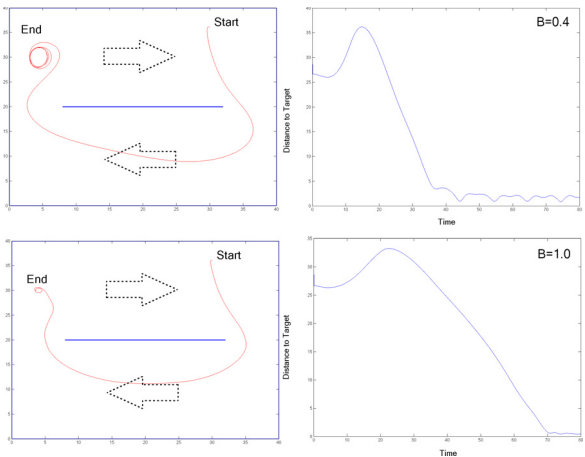

Figure-5: Increasing B reduces transients but slows down motion.

A dampening component that is proportional to velocity exercises omnidirectional attenuation of motion regardless of the direction along which it is heading. This means that the useful component of motion marked by the direction along which the goal component of the gradient of the artificial potential is pointing is treated in the same manner as the unwanted, inertia-induced, noise component of the trajectory. These two components should not be treated equally. Attenuation should be restricted to the inertia-caused, disruptive component of motion, while the component in conformity with the guidance of the artificial potential should be left unaffected (figure-6). A dampening force that takes the above into consideration is:

$$ud = -Bd \cdot [(n^t \dot{X}) n + (\frac{ug^t}{|ug|} \cdot \dot{X} \cdot \Phi(-ug^t \dot{X})) \frac{ug}{|ug|}] \quad (4)$$

where **n** is a unit vector orthogonal to ug, ud represents the dampening force, and Bd is a constant. This force is given the name: nonlinear, anisotropic, dampening force (NADF).

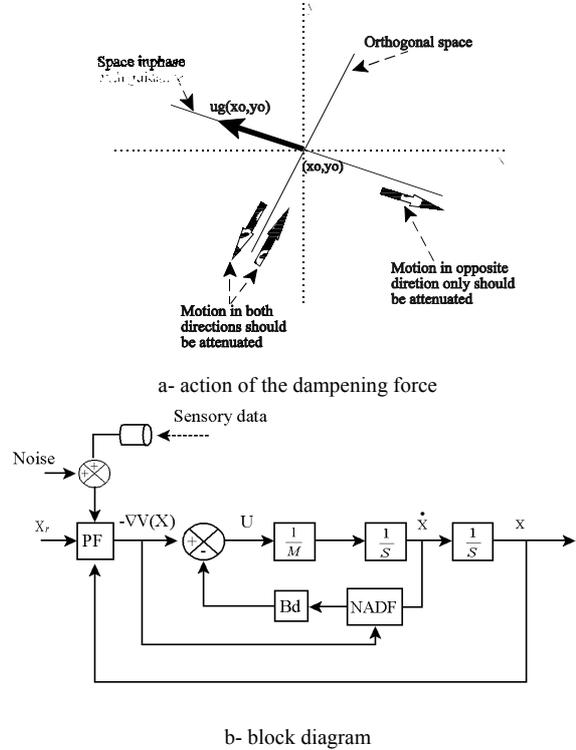

Figure-6: nonlinear, anisotropic, dampening force (NADF).

The NADF coefficient (Bd) is very easy to tune. Since by design the component of motion in conformity with the guidance field is in the null space of NADF, Bd may be set arbitrarily high to attenuate the disruptive competent caused by the robot's inertia. This may be done with no danger of slowing down the robot. The previous example is repeated using NADF. A high Bd of 2.5 is used. The trajectory and convergence curves are shown in figure-7 and the control forces are shown in figure-8. As can be seen the spatial trajectory is well-behaved and a settling time (Ts) of 14 second is obtained. Despite the fact that the coefficient of NADF is two times and a half higher than the linear dampening force coefficient used in figure-5, the system with NADF is more than five times faster.

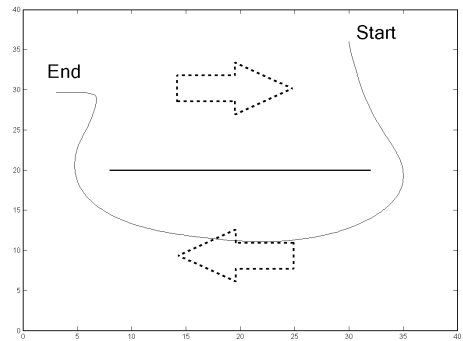

Figure-7a: Trajectory for NADF, Bd=2.5.

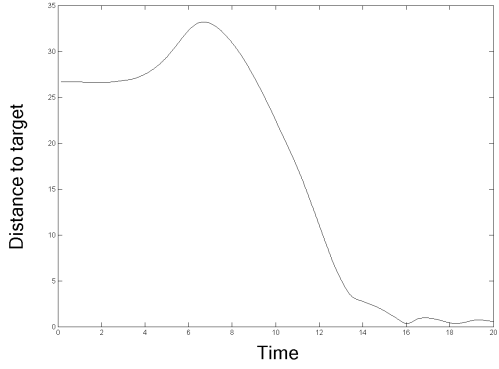

Figure-7b: Distance to target for NADF, Bd=2.5.

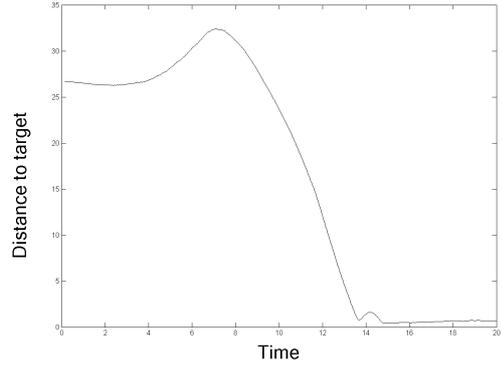

Figure-9b: Distance to the target, sliding mode control.

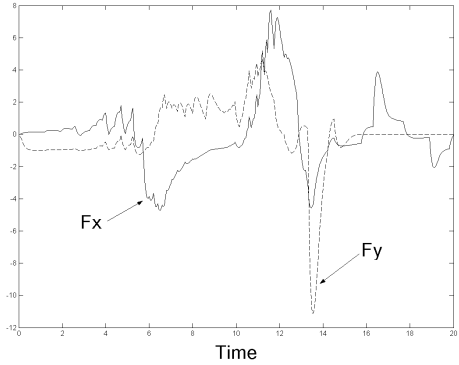

Figure-8: Control signal corresponding to fig. 7.

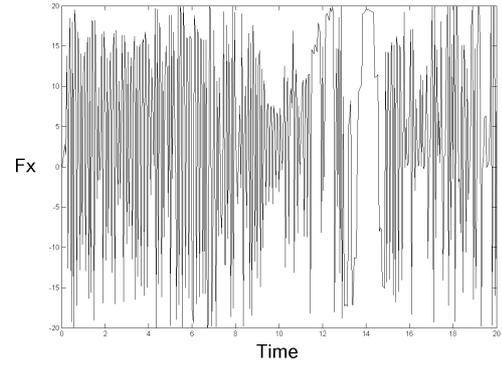

In [20] Guldner and Utkin suggested an approach based on sliding mode (SM) control to convert the gradient guidance field into a control signal. The example in figure-7 is repeated using this approach. The parameters of the SM controllers are adjusted so that the system has a settling time comparable to the one obtained using the NADF approach. Higher level of transients are permitted in order to reduce the magnitude of the control forces the SM controller is exerting. The trajectory and distance to the target versus time are shown in figure-9. The control forces are shown in figure-10. Although the trajectory obtained using the SM control approach has low magnitude high frequency jitters and a higher level of transients near the target compared to the trajectory obtained using the NADF approach, the major difference is in the quality and magnitude of the control signal.

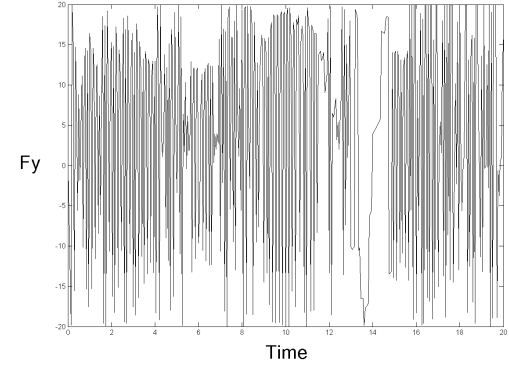

Figure-10: Force control signals, sliding mode control.

### IV. The Nonholonomic Case

The HPF approach has properties that enables it to plan motion for a nonholonomic system. In the following methods are outlined on how to adapt the HPF approach to work with nonholonomic robots to generate both kinematic and dynamic trajectories.

1. Kinematic, Nonholonomic, HPF-based planner:

The equation of motion of a nonholonomic mobile robot may be written as:

$$\begin{bmatrix} \dot{x} \\ \dot{y} \\ \dot{\theta} \end{bmatrix} = \mathbf{G}(\mathbf{x}, \mathbf{y}, \theta, \nu, \omega) \qquad (5)$$

where x and y are the coordinates of the center point of the robot, $\theta$ is its orientation, $\nu$ is the set radial speed of the robot, $\omega$ is the set angular speed, and G is a nonlinear vector function. At a certain (x,y) point in space, equation-5 may be linearized as:

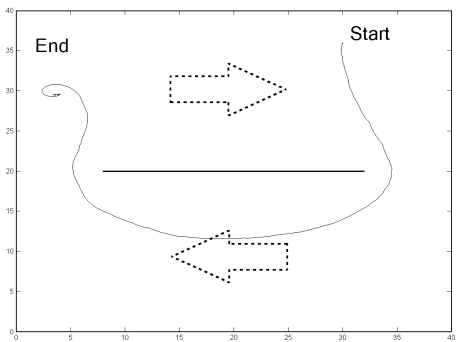

Figure-9a: Trajectory, sliding mode control.

$$\begin{bmatrix} \dot{x} \\ \dot{y} \\ \dot{\theta} \end{bmatrix} = \mathbf{H}(\mathbf{x}, \mathbf{y}, \theta) \begin{bmatrix} v \\ \omega \end{bmatrix}, \qquad (6)$$

where H is a matrix function. The HPF approach can be directly applied to the robot in its linearized form by considering the set radial speed at a certain point in space to be equal to the magnitude of the gradient guidance field at that point and the set angular speed may be taken as the angle between the robot's orientation and the orientation of the gradient guidance field,

$$v = |-\nabla V(x,y)| \qquad (7)$$
$$\omega = \theta - \arg(-\nabla V(x,y))$$

The above procedure can be with little effort adapted to almost any nonholonomic robot. However, in this work we are going to consider planning for a differential drive robot (figure-11).

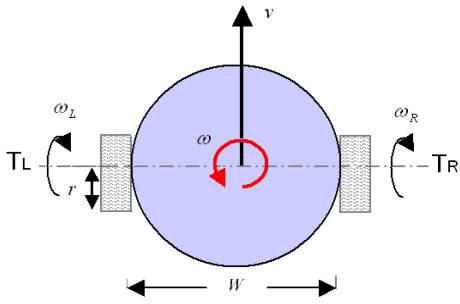

Figure-11: A differential drive mobile robot.

The equations describing motion for such a robot are:

$$\begin{bmatrix} \dot{x} \\ \dot{y} \\ \dot{\theta} \end{bmatrix} = \begin{bmatrix} \cos(\theta) & 0 \\ \sin(\theta) & 0 \\ 0 & 1 \end{bmatrix} \begin{bmatrix} v \\ \omega \end{bmatrix} \qquad (8)$$

and

$$\begin{bmatrix} v \\ \omega \end{bmatrix} = \begin{bmatrix} \dfrac{r}{2} & \dfrac{r}{2} \\ \dfrac{r}{W} & \dfrac{-r}{W} \end{bmatrix} \begin{bmatrix} \omega_R \\ \omega_L \end{bmatrix} = \mathbf{A} \begin{bmatrix} \omega_R \\ \omega_L \end{bmatrix} \qquad (9)$$

where A is the dimension matrix of the robot, r is the radius of the robot's wheels, W is the width of the robot, $\omega_R$ and $\omega_L$ are the angular speeds of the right and left wheels of the robot respectively. The guidance signal derived from the HPF is:

$$\begin{bmatrix} \omega_R \\ \omega_L \end{bmatrix} = \mathbf{A}^+ \begin{bmatrix} |-\nabla V| \\ \theta - \arg(-\nabla V) \end{bmatrix} \qquad (10)$$

where $A^+$ is the pseudo inverse of A. For a differential drive robot $A^+=A^{-1}$. The block diagram of the HPF planner for the kinematic case is shown in figure-12.

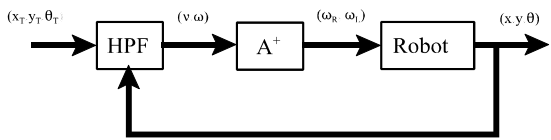

Figure-12: A Kinematic, HPF-based planner, Nonholonomic case.

The above scheme is tested for the gradient guidance field in figure-13. This field encodes the simple behavior of move right and stay at the center of the road (y=0).

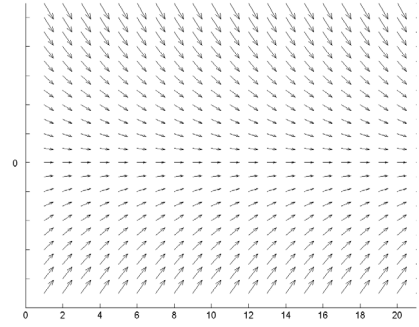

Figure-13: Move right and stay at center gradient guidance field.

The trajectories obtained for different initial orientations are shown in figure-14.

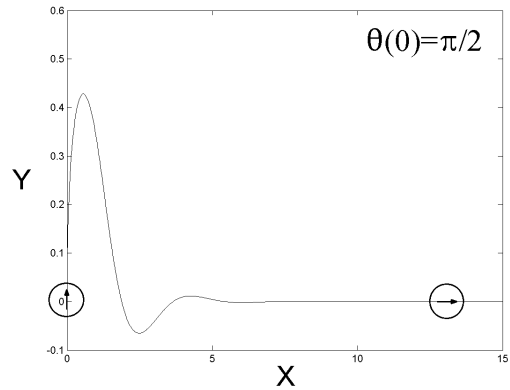

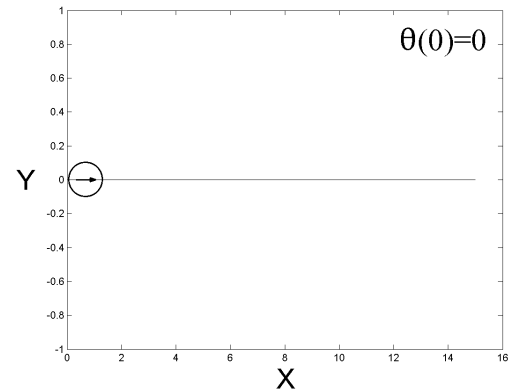

Figure-14: Trajectories from the nonholonomic, kinematic, HPF planner.

2. Dynamic, Nonholonomic, HPF-based planner:

As mentioned earlier, a control signal has to be provided to the robot in order to actuate motion. As demonstrated in the holonomic case, using the kinematic HPF-based planner as the HLC in an HLC-LLC setting may be problematic. Besides the problem of transients a robustness problem may appear. If wheel slip occur, the planner will guide the robot based on false information. In this case problems will arise even if the planner and the controller are functioning properly. While countermeasures against this scenario may be implemented, a planning effort that is less susceptible this type of problems may be derived by making the planner directly dependent on the torques applied to each wheel. If slip occur, the torque of a wheel will drop to zero regardless whether the speed of the wheel changes or not. In this section the idea of NADF is adapted to the nonholonomic case.

The dynamic behavior of the differential drive robot that ties the torques applied to the right and left wheels ($T_R$, $T_L$) to the position and orientation of the robot may be described using two, coupled differential equations. The first one is obtained by differentiating equation-8 with respect to time,

$$\begin{bmatrix} \ddot{x} \\ \ddot{y} \\ \ddot{\theta} \end{bmatrix} = \begin{bmatrix} \cos(\theta) & 0 \\ \sin(\theta) & 0 \\ 0 & 1 \end{bmatrix} \begin{bmatrix} \dot{v} \\ \dot{\omega} \end{bmatrix} + \begin{bmatrix} -\sin(\theta)\dot{\theta} & 0 \\ \cos(\theta)\dot{\theta} & 0 \\ 0 & 0 \end{bmatrix} \begin{bmatrix} v \\ \omega \end{bmatrix}, \quad (11)$$

and the second is derived using Lagrange dynamics in the natural coordinates of the robot,

$$\begin{bmatrix} \dot{v} \\ \dot{\omega} \end{bmatrix} = \frac{1}{M} \begin{bmatrix} \frac{1}{r} & \frac{1}{r} \\ \frac{-4 \cdot r}{W^3} & \frac{4 \cdot r}{W^3} \end{bmatrix} \begin{bmatrix} T_R \\ T_L \end{bmatrix} = \mathbf{B} \cdot \begin{bmatrix} T_R \\ T_L \end{bmatrix} \quad (12)$$

where M is the mass of the robot. Using M=1, the dynamic model of the robot is used instead of the kinematic model in the example shown in figure-14 for the case of $\theta(0)=\pi/2$. As expected direct use of the guidance force as a control signal will fail (figure-15).

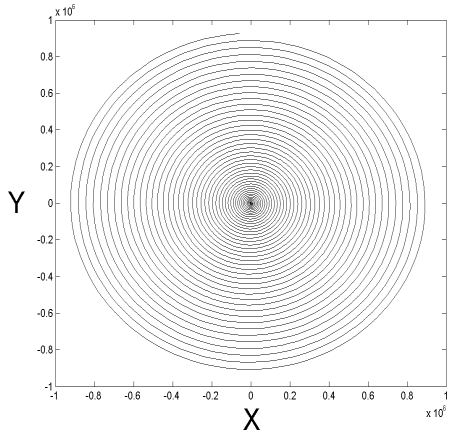

Figure-15: Adding mass causes instability.

To stabilize the system an omni-directional, linear viscous dampening force applied in the natural coordinates of the robot is used to generate the control signal:

$$\begin{bmatrix} T_R \\ T_L \end{bmatrix} = \mathbf{B}^+ \cdot \left[ \mathbf{K}_P \cdot \begin{bmatrix} |-\nabla \mathbf{V}| \\ \theta - \arg(-\nabla \mathbf{V}) \end{bmatrix} - \mathbf{K}_d \cdot \begin{bmatrix} \dot{\rho} \\ \dot{\theta} \end{bmatrix} \right], \quad (13)$$

where $K_P$ and $K_D$ are positive constants, $B^+$ is the pseudo inverse of B, and $\dot{\rho}$ is the radial speed of the robot,

$$\dot{\rho} = \sqrt{\dot{x}^2 + \dot{y}^2} \ . \quad (14)$$

The block diagram of the planner is shown in figure-16.

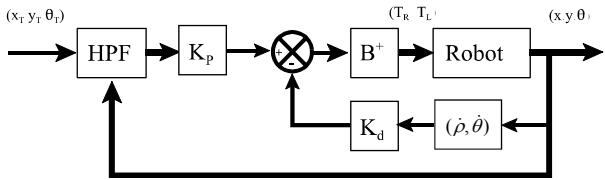

Figure-16: A dynamic, HPF-based planner with linear dampening, nonholonomic case.

The response of the system is shown in figure-17 for different values of $K_P$ and $K_d$. The two cases are simulated for the same duration. As can be seen, the use of rate feed back in the natural coordinates of the robot did stabilize the response and made the system yield to the guidance signal derived from the HPF. Significant transients are observed for a small coefficient of rate feedback. Although increasing this coefficient reduces the transients, it results, as in the holonomic case, in reducing the speed of the robot.

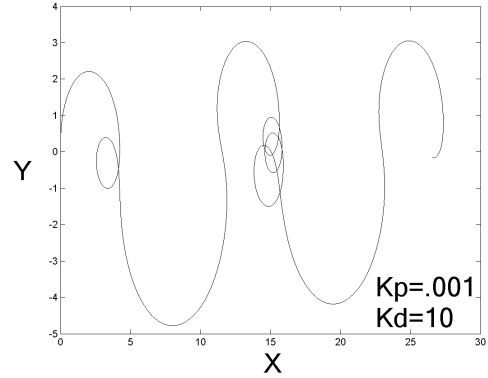

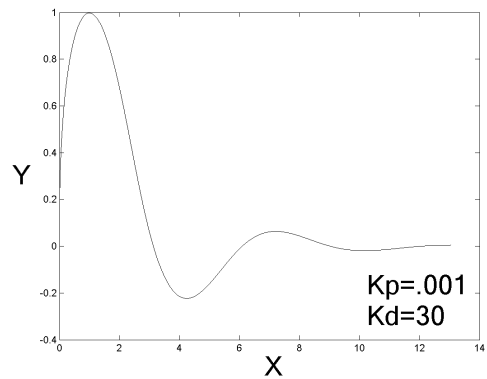

Figure-17: Response of the planner in (13) for different Kp and Kd.

One way to sensitize the dampening to the guidance signal is to notice that changing the speed of the robot is not needed if the actual speed of the system is equal to the reference speed. This leads to a simple, nevertheless effective, change in the form of the control signal:

$$\begin{bmatrix} T_R \\ T_L \end{bmatrix} = \mathbf{B}^+ \cdot \left[ \mathbf{K}_P \cdot \begin{bmatrix} |-\nabla \mathbf{V}| \\ 0 \end{bmatrix} - \mathbf{K}_d \cdot \begin{bmatrix} \dot{\rho} \\ \dot{\theta} - (\theta - \arg(-\nabla \mathbf{V})) \end{bmatrix} \right] \quad (15)$$

In figure-18, the direction sensitive dampening is compared to the linear dampening case using same coefficients for the planner. As can be seen sensitizing the dampening to direction significantly reduced the overshoot and settling time without compromising the speed of the robot.

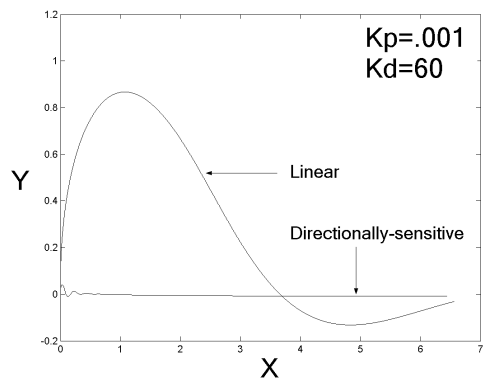

Figure-18: response of the planner in (15) compared to the one in (13).

The performance can still be further enhanced by making the reference radial speed at a certain point dependant on the orientation of the robot relative to the orientation of the guidance vector. The reasoning that may be used is: if the two orientations are the same use maximum reference speed. If the two orientations are at right angle use zero reference speed, and if the two orientations are diametrically opposite use a negative maximum reference speed. This reasoning may be implemented by simply multiplying the reference speed with cosine the difference between the two orientations. The control signal that realizes the above is:

$$\begin{bmatrix} T_R \\ T_L \end{bmatrix} = B^+ \cdot \left[ K_P \cdot \begin{bmatrix} |-\nabla V| \cdot \cos(\theta - \arg(-\nabla V)) \\ 0 \end{bmatrix} - K_d \cdot \begin{bmatrix} \dot{\rho} \\ \dot{\theta} - (\theta - \arg(-\nabla V)) \end{bmatrix} \right] \quad (16)$$

In figure-19 the direction sensitive controller in (15) is compared to the jointly sensitized controller in (16). As can be seen the jointly sensitive controller lead to more reduction in the overshoot.

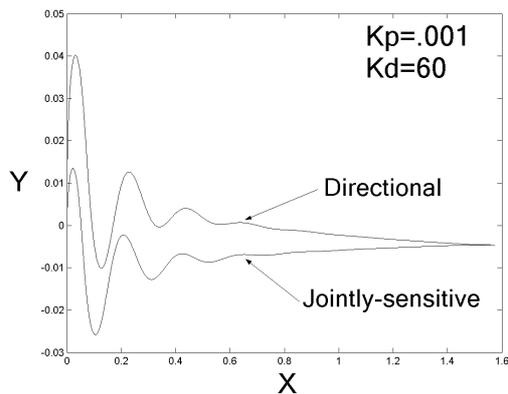

Figure-19: response of the planner in (15) compared to the one in (16).

The ability of the controller to track a unit square pulse and a sinusoid are shown in figure-20.

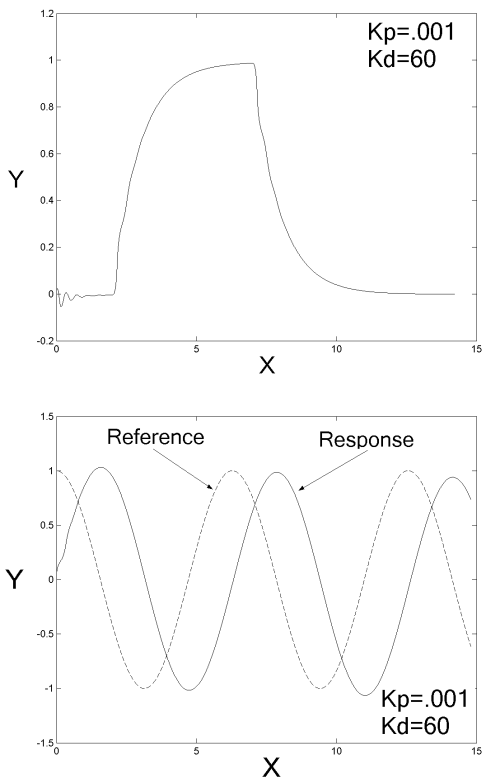

Figure-20: square pulse and sinusoidal tracking ability of the planner in (16).

Using a $K_p$=.001 and a $K_d$=60, The controller in (16) is tested in a cluttered environment. Figure-21 shows the harmonic gradient guidance field that is used to motivate the motion of the robot and the holonomic, kinematic trajectory such a field generates. Figure-22 shows the dynamic trajectory the controller generates and the orientation of the robot as a function of time. As can be seen, the nonholonomic, dynamic trajectory is very close in shape to the holonomic, kinematic trajectory with a satisfactorily smooth orientation profile. The control torques applied to the right and left wheels of the robot are shown in figure-23.

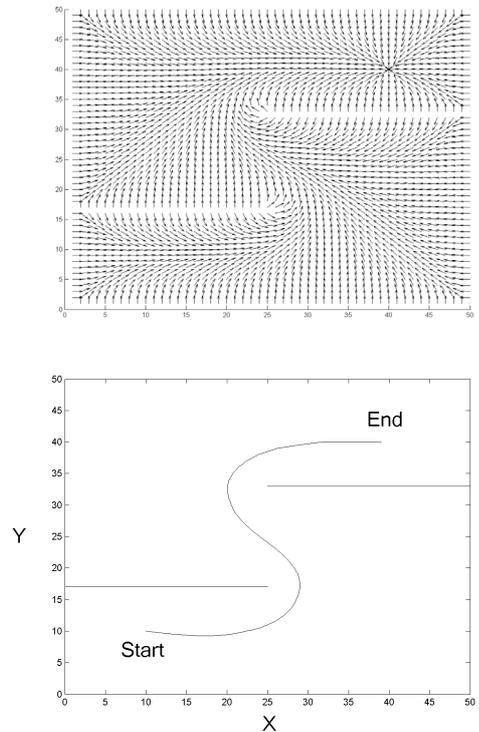

Figure-21: Guidance field and trajectory of a kinematic, holonomic, HPF planner.

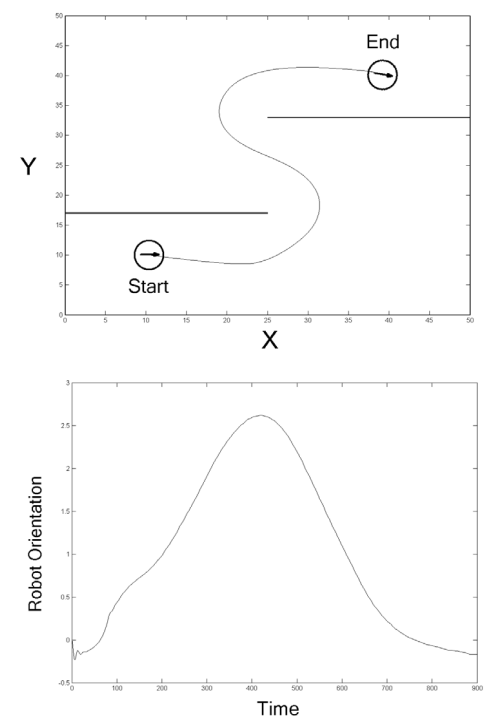

Figure-22: Trajectory and curvature using the planner in (16) and the guidance field in fig. 21.

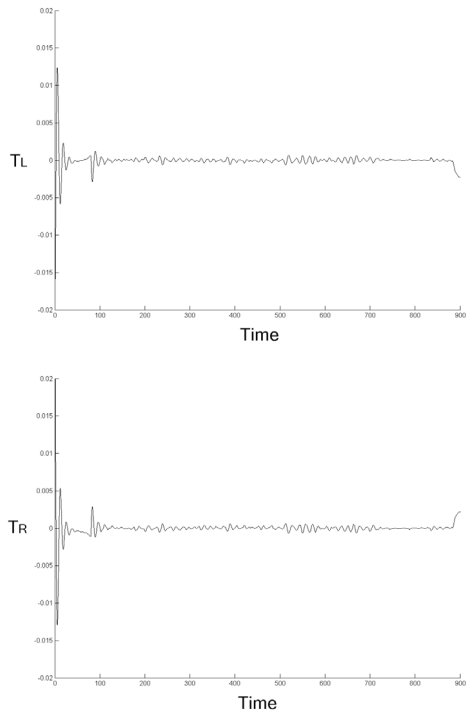

Figure-23: Torque control signals corresponding to fig. 22.

## IV. Conclusions

In this paper the harmonic potential field motion planning method is cast in a navigation control framework where *a priori* data about a situation is directly converted into a control signal. The gradient of a harmonic potential field, which can only provide a guiding reference, is converted into a control signal using the NADF concept suggested in this paper. As was demonstrated, attempting to convert the gradient field into a control signal by adding a linear viscous dampening force (a force proportional to velocity) may be problematic. On the other hand, carrying out such an extension using the NADF approach is straightforward and practical. This is because the NADF approach is developed to take into consideration the dual role the gradient field of an HPF plays both as a control signal and a guidance provider. The simultaneous consideration of these two factors is what enables the control signal to effectively suppress transients without slowing down motion. The work in this paper may be considered as another step towards the HPF approach attaining its full potential.

**Acknowledgment**
The author acknowledges KFUPM support of this work.